\documentclass[letterpaper, 10 pt, conference]{ieeeconf}  
\pdfoutput=1
\usepackage{pdfpages}

\IEEEoverridecommandlockouts                              
\usepackage{graphics} 
\usepackage{epsfig} 
\usepackage{mathptmx} 
\usepackage{times} 
\usepackage{amsmath} 
\usepackage{amssymb}  
\usepackage{verbatim}
\usepackage{csquotes}
\usepackage{graphicx}
\usepackage{cite}
\usepackage{amsmath}
\usepackage{booktabs}
\usepackage{diagbox}
\usepackage{multirow}
\usepackage{commath}
\usepackage{relsize}
\usepackage{subcaption}
\usepackage[keeplastbox]{flushend}
\usepackage{gensymb}

\title{\LARGE \bf
MAVI: A Research Platform for Telepresence and Teleoperation\\
{\small
Technical Report, May 2018
}
}

\author{Mojtaba Karimi, Tamay Aykut, and Eckehard Steinbach
\thanks{All authors are with the Department of Electrical and Computer Engineering, Technical University of Munich, Germany
        {\tt\small \{mojtaba.karimi, tamay.aykut, eckehard.steinbach\}@tum.de}}%
}

\begin{document}
\maketitle
\thispagestyle{empty}
\pagestyle{empty}

\begin{abstract}
One of the goals in telepresence is to be able to perform daily tasks remotely. A key requirement for this is a robust and reliable mobile robotic platform. Ideally, such a platform should support 360$\degree$ stereoscopic vision and semi-autonomous telemanipulation ability. In this technical report, we present our latest work on designing the telepresence mobile robot platform called MAVI.
MAVI is a low-cost and robust but extendable platform for research and educational purpose, especially for machine vision and human interaction in telepresence setups. The MAVI platform offers a balance between modularity, capabilities, accessibility, cost and an open source software framework. With a range of different sensors such as Inertial Measurement Unit (IMU), 360$\degree$ laser rangefinder, ultrasonic proximity sensors, and force sensors along with smart actuation in omnidirectional holonomic locomotion, high load cylindrical manipulator, and actuated stereoscopic Pan-Tilt-Roll Unit (PTRU), not only MAVI can provide the basic feedbacks from its surroundings, but also can interact within the remote environment in multiple ways. The software architecture of MAVI is based on the Robot Operating System (ROS) which allows for the easy integration of the state-of-the-art software packages.
\end{abstract}

\section{INTRODUCTION}\label{sec:introduction}
Telepresence systems allows users to perform various tasks in a remote environment \cite{tachi2016telexistence}. This functionality supports a broad range of applications, such as teleconferencing \cite{neustaedter2016beam}, remote healthcare \cite{anton2017augmented, koceski2016evaluation}, and inspection in hazardous environments \cite{lombard15,jakuba2016nereid}. Teleoperated mobile robots are used to perform complex tasks in remote environments and can be considered as a particular example of telemanipulation systems \cite{Piltan15}. One challenge is to have a low-cost mobile robot with manipulation ability that can be easily operated semi-autonomously to perform simple tasks in the remote environment. 
\begin{figure}[ht]
    \vspace{-0.5em}
    \centering
    \includegraphics[width=0.5\textwidth]{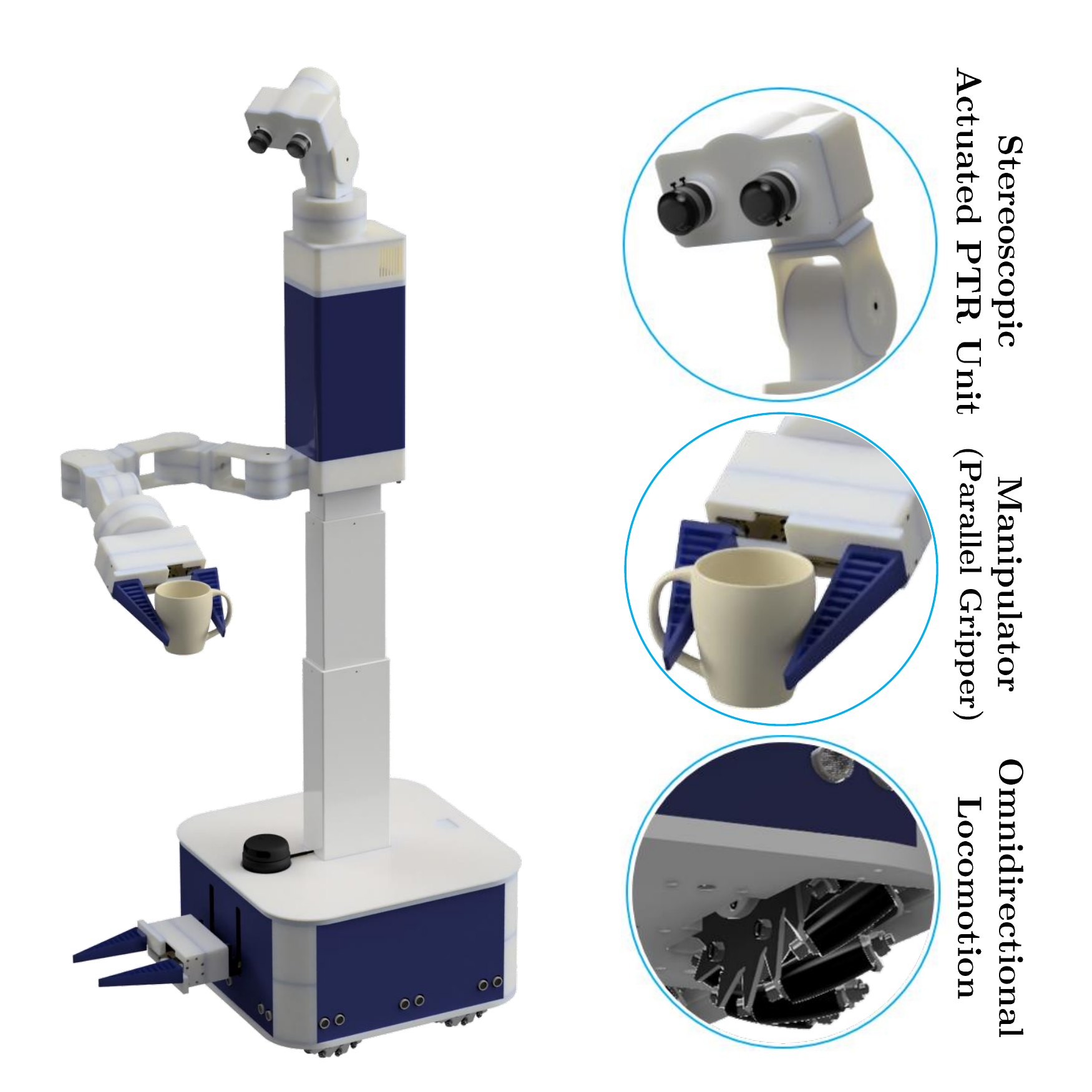}
    \caption{{\small The MAVI telepresence platform consists of three main parts: an actuated stereoscopic Pan-Tilt-Roll Unit (PTRU), a cylindrical manipulator with parallel gripper, and an omnidirectional holonomic locomotion system.}}
    \label{fig:MAVI}\vspace{-1.80em}
\end{figure}

In this regard, telepresence has been applied on different robotic platforms, mostly with simple locomotion systems equipped with a monocular camera and a screen as a human-machine interface \cite{laniel2017enhancing, zhang2018a360}. Although these platforms provide the basic functionality for telepresence applications, they incorporate limited human morphology and mostly do not consider bidirectional user interaction. In particular, the usage of a single camera for capturing the remote environment provides only a limited illusion of spatial attendance. Furthermore, classic motion control of mobile platforms in today’s telepresence systems are often restricted to simple interaction devices. Accordingly, an effective manipulation capability and proficient locomotion system along with actuated camera systems are all equally important to provide a better immersion and telepresence experience \cite{Jon92}. 

In this paper, we introduce the MAVI telepresence platform which is designed specifically for semi-autonomous indoor teleoperation scenarios. MAVI is a low-cost telepresence robot with omnidirectional holonomic locomotion, a cylindrical 7DoF manipulator, and an actuated stereoscopic vision system. This robot is designed to perform a variety of remote mobile manipulation tasks in an everyday environment. Our design can simply execute the basic and complex commands of a human user in an indoor area and provides most of the sensing modalities such as visual and force feedback placed in different parts of the platform.
\begin{figure*}[!ht]
    \centering
    \includegraphics[width=\textwidth]{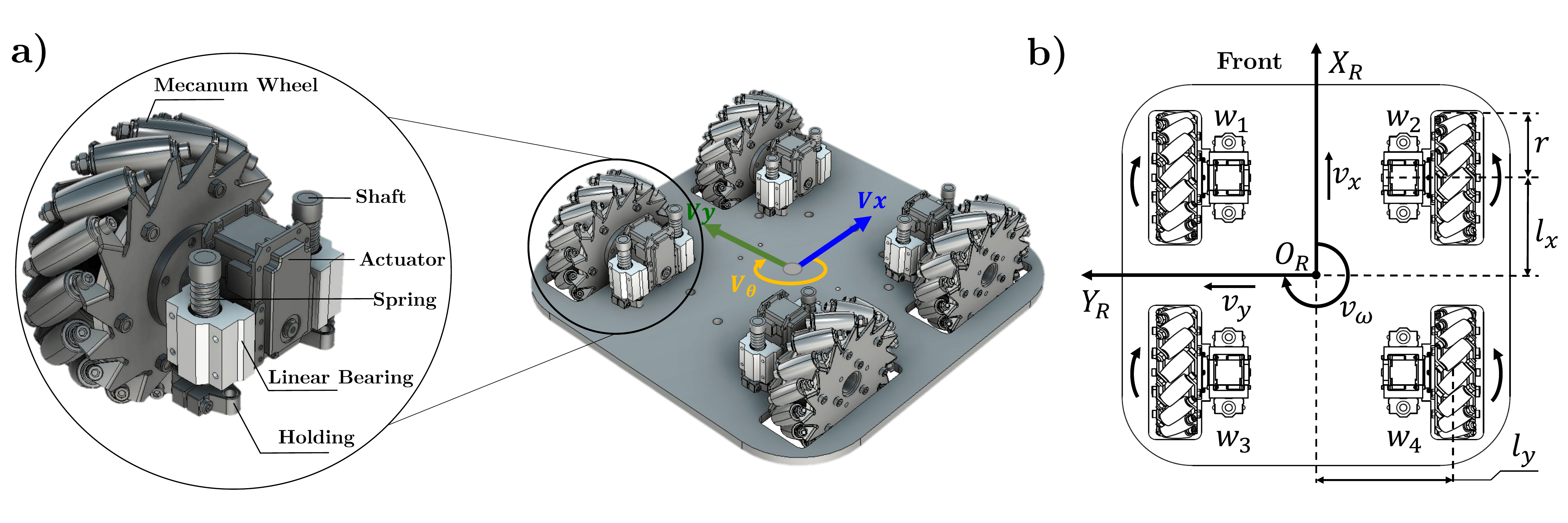}
    \caption{{\small MAVI omnidirectional holonomic locomotion system overview. (a) shows the low-cost linear vertical suspension system designed for Mecanum wheels and, (b) illustrates the coordinate system of the locomotion system with design dimensions in order to calculate the IK and odometery of the robot.}}
    \label{fig:locomotion}\vspace{-0.01em}
\end{figure*}
The structure of this report is as follows: In Section~\ref{sec:telepresence} an overview of the proposed telepresence system is given. Then, in Section~\ref{sec:mavi}, the MAVI mobile robot hardware architecture is explained. In Section~\ref{sec:software}, the software framework is discussed and the structure of the node based system is shown. Also, the control packages on the robot side are illustrated. Section~\ref{sec:conclusion} concludes with final remarks.

\section{TELEPRESENCE SYSTEM OVERVIEW}\label{sec:telepresence}
The proposed telepresence system consisted of a client station (User) and the server station (Robot), both remotely located and connected to the communication network. The user can control the MAVI platform in an indoor environment by sending the commands from the client side. The software framework for the MAVI platform is based on ROS (Robot Operation System), with the transmission of the data between the robot and the client being defined as ROS based messages \cite{quigley2016ros}.

\paragraph*{Server Station (Robot)}
The MAVI platform design is based on three main units: omnidirectional locomotion system, cylindrical manipulator, and stereoscopic Pan-Tilt-Roll Unit (PTRU). The robot is able to connect to the 2.4 or 5.0GHz wireless network to transmit and receive the visual, auditory, haptic and motion data. In order to make the teleoperation as accurate and simple as possible, high-level commands are transmitted to control the manipulator and locomotion system while the local control loop on the robot side generates appropriate actuation signals. An actuated PTRU with high-resolution camera can provides a stereoscopic 360$\degree$ view of the environment and can mimic the user's head motion. MAVI is equipped with a variety of sensors to estimate its state within the indoor environment and provide high-level feedback to be easily operated. The MAVI locomotion system is an omnidirectional holonomic system which allows the user to easily maneuver. The manipulator ability of the platform supports simple picking and placing of objects by its large payload and working-space.

\paragraph*{Client Station (User)}
On the user side, our system acquires and sends the high-level commands for locomotion, manipulation, and head motion as well as provides video, audio, and haptic feedback to the user. In order to simplify the user interface to control the robot, we are using a gesture armband from Wearable Technology by Thalmic labs \cite{myo2018}. This armband contains two types of sensors: high rate Inertial Measurement Unit (IMU) and EMG sensors for capturing the muscle activities of the arm and the fingers. Generating commands according to the user's arm orientation and their gestures allows us to control both locomotion and the manipulation system of the MAVI platform.
To provide video and audio feedback, the client side is equipped with a Head-Mounted-Display (HMD) combined with headphones to show the stereo video and play the audio signals. To provide haptic feedback, we are currently using the vibration ability of the armband. An internal IMU of the HMD is used to estimate the users head motion which is sent to the robot side to control the PTRU.

\section{MAVI ARCHITECTURE}\label{sec:mavi}
As shown in Fig.~\ref{fig:MAVI}, MAVI consists of a human-size torso and a mobile base with 1.65 m height and 0.42 m in both width and depth. The platform is designed for an indoor environment and a working time of 4 to 8 hours depending on the different scenarios. MAVI is constructed mostly from light-weight 3D-printed parts. Robotis Dynamixel actuators drive all joints in the robot manipulator, the PTRU, and the locomotion system \cite{mx106}. This allows for a light-weight and low-cost construction, compared to available telepresence robots. The MAVI hardware structure is divided into the following parts:

\paragraph*{Holonomic Locomotion}
Omnidirectional locomotion of MAVI is ensured by a Mecanum drive mechanism to maneuver in the narrow passages found in indoor areas. We developed an omnidirectional driving system based on four Mecanum wheels equipped with a suspension system to be able to smoothly bridge irregularities at the ground (see Fig.~\ref{fig:locomotion}). An omnidirectional locomotion system based on four Mecanum wheels, compared to other traditional holonomic systems, can carry more weight and provides more efficient movement \cite{ishigami2015design, Hen96}. On the other hand, one of the problems for this structure is the gap between each freewheel which causes a periodic vibration in the robot. Also critical are bumps or puddles on the ground, which not all the wheels may touch the ground plane to generate the desired movement, as a result, driving of the robot cannot be controlled appropriately \cite{ishigami2015design}. 

To address these issues, all the wheels need to be connected to a suspension system to not only absorb the vibration while the robot is moving but also to push all the wheels to touch the ground. Fig.~\ref{fig:locomotion}(a), shows the structure of the wheels actuation mechanism with the linear vertical suspension system. The Mecanum wheel and the actuator are assembled as one unit and connected to two cylinder slides trough linear bearings from sides, while the absorber spring placed in between the body plate and the linear bearings.

\begin{figure*}[!ht]
    \centering
    \includegraphics[width=\textwidth]{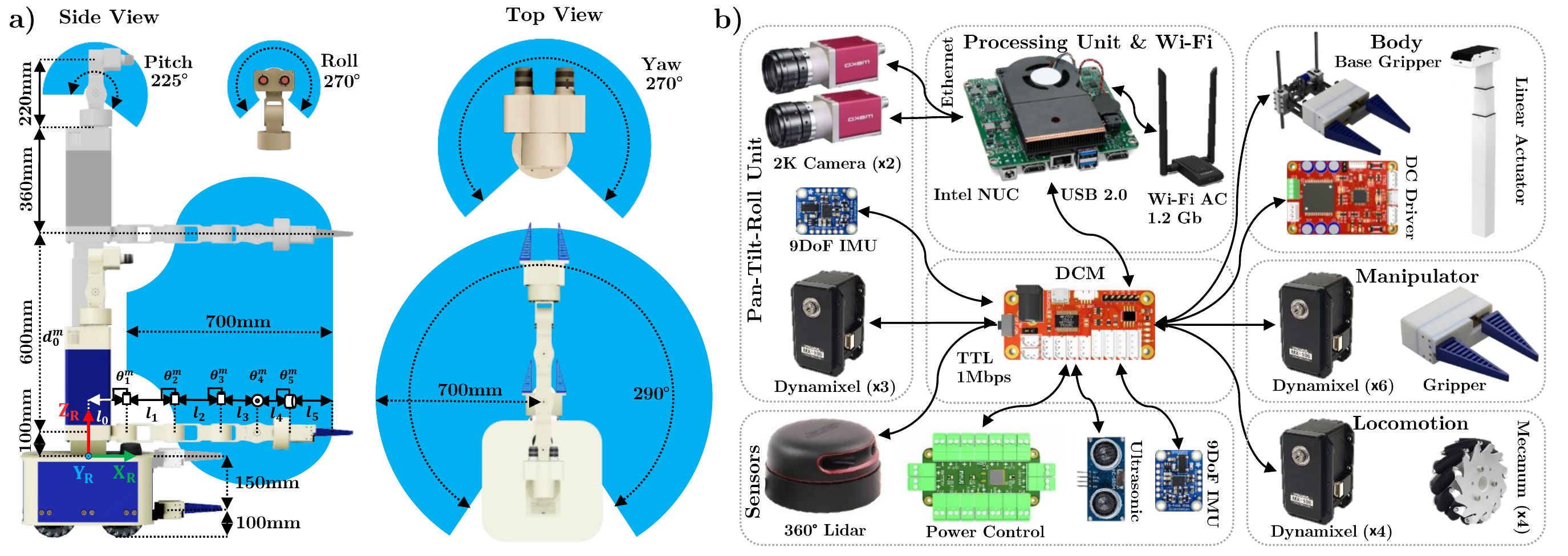}
    \vspace{-0.8em}
    \caption{{\small a) The MAVI's torso dimension and it's cylindrical manipulator and Pan-Tilt-Roll Unit (PTRU) workspace. b) Electronic system overview of the platform. All the sensors and the actuators are communicate to the main processor using a id based Half-Duplex serial TTL bus at 1.0Mbps.}}
    \label{fig:man@elec}\vspace{-1.01em}
\end{figure*}

To control the velocity of the robot on a planner surface, the inverse kinematic of the locomotion system needs to be calculated \cite{tatar2014design}. Fig.~\ref{fig:locomotion}(b), shows the schematic of the designed system for the platform and illustrates the axes of the locomotion system. By considering the design parameters, the inverse kinematic of the robot locomotion is defined as follow: \begin{equation}
\begin{split}
\omega_1 &= 1/r (\text{V}_x + \text{V}_y - ((\text{l}_x + \text{l}_y)-\text{V}_\omega ))\\
\omega_2 &= -1/r (\text{V}_x - \text{V}_y + ((\text{l}_x + \text{l}_y) -\text{V}_\omega ))\\
\omega_3 &= 1/r (\text{V}_x - \text{V}_y - ((\text{l}_x + \text{l}_y) -\text{V}_\omega ))\\
\omega_4 &= -1/r (\text{V}_x + \text{V}_y + ((\text{l}_x + \text{l}_y) -\text{V}_\omega ))\\
\end{split}
\end{equation}
where $\text{V}_x$ and $\text{V}_y$ are the longitudinal and the lateral velocities, and $\text{V}_\omega$ is the angular velocity of the mobile robot. The rotating speed of the i-th wheel is $\omega_i$ in rad/s. In addition, $\text{l}_x$ and $\text{l}_y$ are the distance between the wheels in longitudinal and lateral direction respectively, and $r$ is the radius of the Mecanum wheel. Meanwhile, according to the rotation feedback from the internal shaft encoder of each wheel $\omega_i^s$ , we calculate the odometery of the robot as follows:
\begin{equation}
\begin{split}
P_x^\text{odom} &= \sum \limits_{i=0}^t \frac{r}{4} (\omega_1^s + \omega_2^s + \omega_3^s + \omega_4^s ) dt\\
P_y^\text{odom} &= \sum \limits_{i=0}^t \frac{r}{4} (-\omega_1^s + \omega_2^s + \omega_3^s - \omega_4^s ) dt\\
P_w^\text{odom} &= \sum \limits_{i=0}^t \frac{r}{4(l_x+l_y)} (-\omega_1^s + \omega_2^s - \omega_3^s + \omega_4^s ) dt\\
\end{split}
\end{equation}
where $dt$ is the time elapsed, $P_x^\text{odom}$ and $P_y^\text{odom}$ are the positions of the robot in meter along with longitude and latitude, respectively, and $P_\omega^\text{odom}$ is the orientation of the robot in radian relative to the starting point.\\
\vspace{0.1em}
\begin{figure}[ht]
    \centering
    \includegraphics[width=0.49\textwidth]{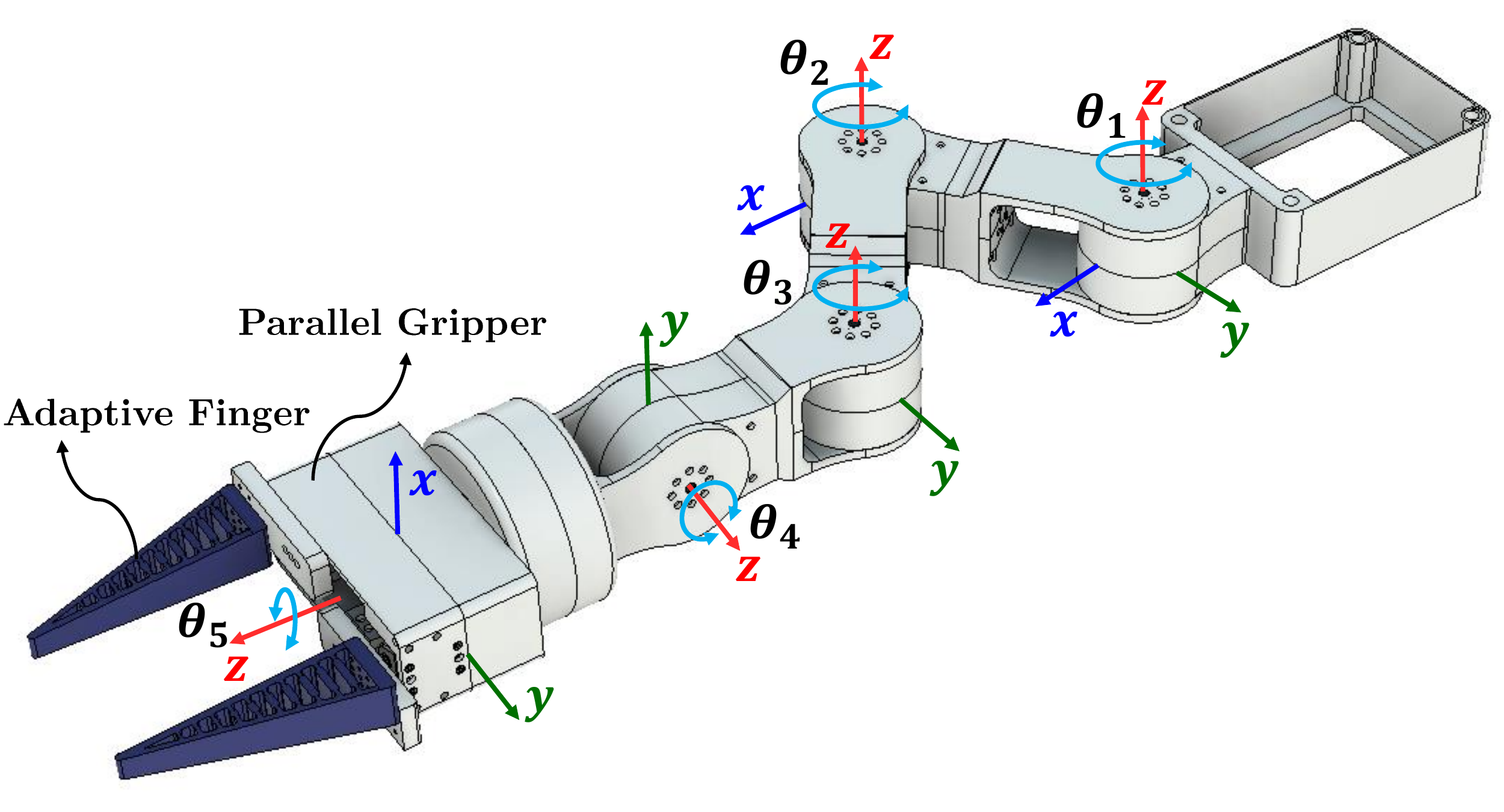}
    \vspace{0.3em}
    \caption{{\small Planner 6DoF manipulator design characteristics. This structure grantees high payload (up to 0.8kg) and large workespace while it is using low-cost actuators. An adaptive finger from FESTO is used for parallel gripper design\cite{festo120}.}}
    \label{fig:mansch}\vspace{-0.8em}
\end{figure}
\vspace{-0.2em}
\paragraph*{Cylindrical Manipulator}
The high payload capability of the MAVI manipulator is provided by its particular design. It is a 7DoF cylindrical manipulator, consisting of a vertical linear actuated body mounted on a locomotion base, a 3DoF planar portion followed by a 2DoF wrist (pitch and roll) and equipped with a parallel gripper facilitating a payload of 1.0kg. The planar part of the manipulator provides a mechanical support and increases the amount of force that can applied to the end-effector given the large lever arm (see Fig.~\ref{fig:mansch}). The workspace of the manipulator is shown in Fig.~\ref{fig:man@elec}(a). It is slightly constrained but can be compensated by the robots mobility. We further added another movable gripper to the base to avoid complex motion plannings for trivial obstacle removal or pick and place tasks. The resulting joint manipulation capabilities enable a performant task accomplishment.

The anthropomorphic manipulation control by the user can be realized by mapping the user's arm gesture, e.g. using the gesture band in \cite{myo2018}, to 6D control commands. To simultaneously control the manipulator and provide real-time force feedback, we implemented a compliance control for the joints\cite{quigley2011low}. This functionality is enabled by the joints with back-drivable servo actuators. Rather than using a time expensive inverse kinematic (IK) solver to control the custom-designed MAVI manipulator, we resolved the IK problem algebraically. We can determine the revolute and prismatic joint parameters of the manipulator, $\theta_{1:5}^m$ and $d_0^m$ respectively, from a desired pose command of the user ($X, Y, Z, \theta, \phi, \psi$) with $l_{1:5}$ which are the length of the joint links (see Fig.~\ref{fig:mansch} and Fig.~\ref{fig:man@elec}(a)).
\begin{equation}
\begin{split}
x &=X-l_0-(l_3+(l_4+l_5)\scalebox{1.0}{\text{cos}}(\theta))\scalebox{0.9}{\text{cos}}(\phi)\\
y &=Y-(l_3+(l_4+l_5)\scalebox{1}{\text{cos}}(\theta))\scalebox{0.9}{\text{sin}}(\phi)\\
\theta_1^m &=\scalebox{1.0}{\text{atan}}(y,x)-\scalebox{1.0}{\text{acos}}((x^2+y^2+l_1^2+l_2^2)/(2l_1\sqrt{x^2+y^2}))\\
\theta_2^m &=\pi-\scalebox{1.0}{\text{acos}}{(({-x^2-y^2+l_1^2+l_2^2)/(2 l_1 l_2}))}\\
\theta_3^m &=\psi-\theta_1^m-\theta_2^m\\
\theta_4^m &=\theta\\
\theta_5^m &=\psi\\
d_0^m &=Z-(l_4+l_5){\scalebox{1.0}{\text{sin}}}(\theta)\vspace{-.9em}
\end{split}
\end{equation}
where $x$ and $y$ are the temporary position for the end-effector and used to calculate the final joint orientations.
\vspace{-0.05em}
\paragraph*{Actuated PTRU}
A stereo camera system is mounted on a 3DoF actuated PTRU and follows the yaw, pitch, and roll (ZYX) Euler convention with a large working space as shown in Fig.~\ref{fig:man@elec}(a). The distance between the cameras is adjustable from 60mm about +/-10mm according to the user. Any desired head orientation triggered by the user is converted to quaternions. The correct conversion of an arbitrary head orientation $q$ to the angles $\theta_c$, $\phi_c$, and $\psi_c$ subject to our custom PTRU design can be obtained with the following equation:
\vspace{-0.4em}
\begin{equation}
\begin{split}
\theta_c &= \scalebox{0.9}{\text{atan2}}(-2(q_\text{x} q_\text{y} - q_\text{w} q_\text{z}), q_\text{w}^2+q_\text{x}^2-q_\text{y}^2-q_\text{z}^2))\\
\phi_c &= \scalebox{0.9}{\text{asin}}(2(q_\text{x} q_\text{z} + q_\text{w} q_\text{y}))\\
\psi_c &= \scalebox{0.9}{\text{atan2}}(-2(q_\text{y} q_\text{z} - q_\text{w} q_\text{x}), q_\text{w}^2-q_\text{x}^2-q_\text{y}^2+q_\text{z}^2)
\vspace{-0.4em}
\end{split}
\end{equation}
\vspace{-0.5em}
\paragraph*{Processing and Sensing}
All the control boards (actuators and the sensor drivers) are designed as a distributed module \cite{karimi2015remoro}. Each module connected to a system serial bus and communicates over Half-Duplex Serial TTL at 1.0Mbps.
MAVI perceives its environment with a variety of sensors. The robot senses the surrounding with a 360$\degree$ Lidar (RP-LIDAR A2) \cite{huang18}, along with twelve ultrasonic rangefinders. They are mainly used for 2D mapping and localization as well as obstacle avoidance and providing a virtual feedback of the obstacle proximity on the user side. MAVI is also equipped with two 9-Axis inertial motion sensors (BNO055 IMU) \cite{bno055}, one mounted on the body frame and the other on the PTRU (next to the stereo camera).
Each IMU sensor provides filtered absolute orientation, linear acceleration, and the rotation speed for all three axes and up to 100Hz.

The main processing unit on MAVI is an Intel NUC, a powerful 4x4-inch mini PC with Core-i7 CPU and 32GB of RAM \cite{intelnuc}. Two stereo cameras with fish-eye lens are connected through a 1.0Gbps Ethernet switch to the processing unit. The other additional modules like microphone and speaker are connected to the main processor's sound card.
\section{SOFTWARE FRAMEWORK}\label{sec:software}
The software architecture of the MAVI platform is based on the Robot Operating System (ROS) \cite{quigley2016ros}. ROS is an open source framework, including different tools, packages and libraries that make it possible to run and implement different robotic components to communicate with each other. Individual programs, also called nodes, communicate over topics which is a name for a stream of messages on the system. A ROS message consists of a nested structure of variables or objects. The ROS framework also enables the communication between multiple computers running in the same network. This ability of the ROS framework allow us to access all the sensory data from the server station. This means only critical real-time nodes needs to run on the robot's internal processor while the other non real-time control nodes can be executed in the client station (user side).
Fig.~\ref{fig:software} shows the structure of the nodes for the MAVI telepresence platform. In the first layer (Driver Layer), a Device Communication Manager (DCM), which implemented as a ROS package, is running at 100Hz reading all the sensory data in real-time. Nodes in this package communicate to all the devices which connected to the serial bus in the platform. As shown in Fig.~\ref{fig:man@elec}(b), all the sensors and actuators are connected to the serial bus to communicate with DCM. Each module in the bus has a unique id address and this structure allows us to easily add or remove different hardware modules from the platform. In the middle layer (Control Layer) of the software framework, MAVI has four main packages: 
\begin{itemize}\setlength\itemsep{0.1em}
\item
\textit{Locomotion Controller}: Reads the sensory information from the wheels (encoder, voltage, current, and force) along with body IMU data and controls the omnidirectional movement of the robot.
\begin{figure}[ht]
    \centering
    \includegraphics[width=0.49\textwidth]{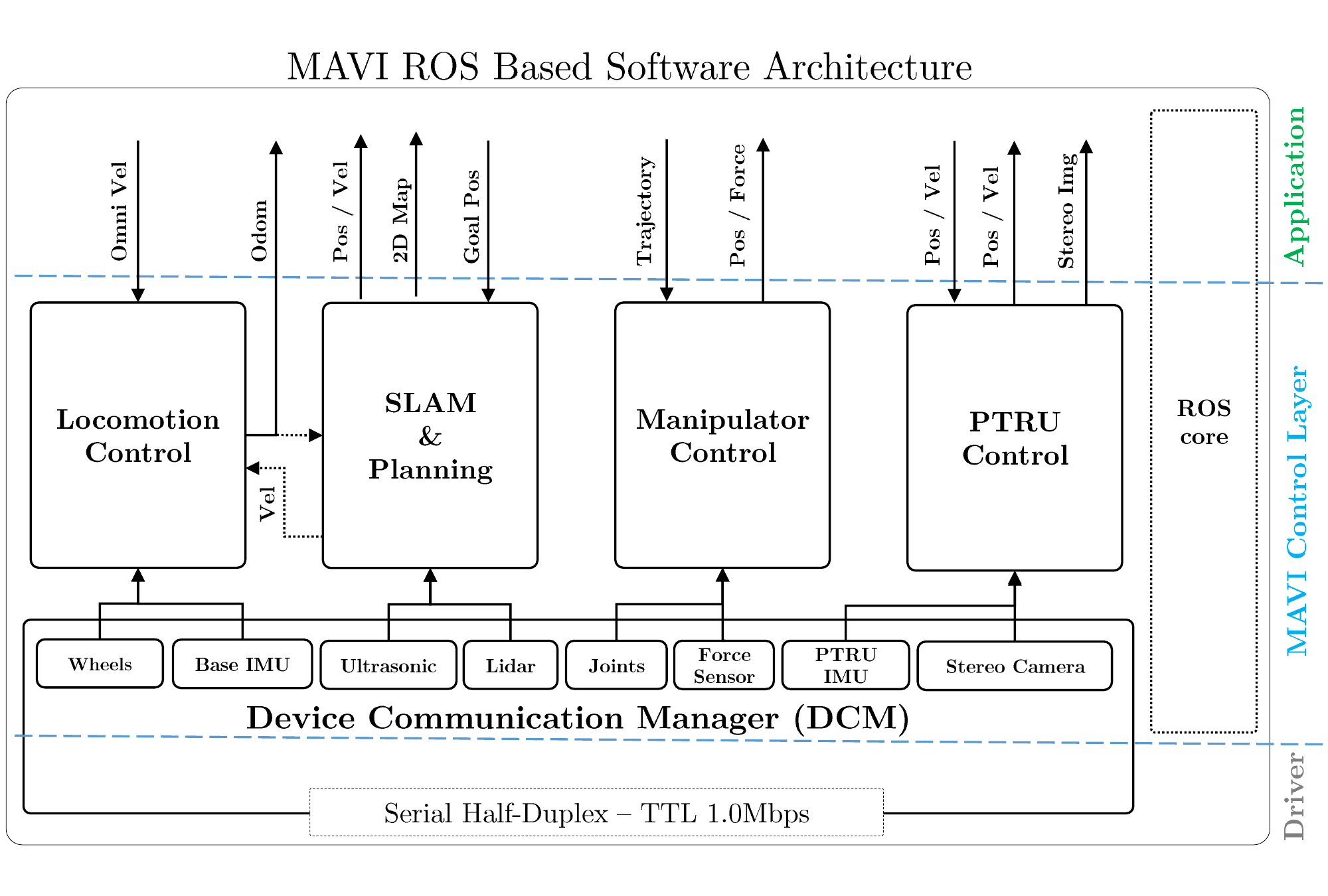}
    \vspace{-1.8em}
    \caption{{\small A ROS based software framework architecture for the MAVI platform. Driver layer is a hard real-time unit to communicate to the distributed modules. The control layer runs the local real-time loops and communicate to the application layer.}}
    \label{fig:software}\vspace{-1.8em}
\end{figure}
\item
\textit{Manipulator Controller}: To control the cylindrical manipulator for given trajectory poses. This unit runs at 100Hz and it uses all the sensory data from the manipulator joints (such as: position, speed, force, current, temperature and voltage) to control the manipulator and the parallel gripper.
\item
\textit{PTRU Controller}: This package uses all the sensory data from the actuators of the PTRU and also the head IMU data and controls the motion of the PTRU actuators at 100Hz. We are using a delay compensation strategy for actuated stereoscopic vision system based on head motion prediction \cite{TA17, TA18a}. Publishing the stereo camera information is also integrated in this package and runs at 30Hz.    
\item
\textit{Localization and Mapping}: This unit uses a laser range finder, odometery output, ultrasonic range finder and the body IMU data for SLAM. It utilizes Gmapping package to provide a map and the position of the robot in an indoor environment \cite{grisetti2007improved}. This control loop is running under 10Hz and provides the path planning and a 2D map of the indoor environment.
\end{itemize}
\vspace{-0.1em}
\section{CONCLUSIONS}\label{sec:conclusion}
In this report, we present our low-cost MAVI telepresence robot platform and provide a general system description. The proposed telepresence omnidirectional robot is specifically designed for an indoor tele-operation scenario. While typical telepresence robots mostly can not interact within the environment in a multimodal way, we introduce our novel design of a cylindrical manipulator with large payload and customized working space. It is also equipped with the parallel gripper with adaptive fingers to provide easy and reliable remote manipulation. To increase the reliability and performance of the holonomic locomotion system, we adapt the simple four wheel Mecanum drive structure with suspension mechanism. To provide better immersion in the remote environment, we integrate a stereoscopic vision system with actuated Pan-Tilt-Roll Unit. Finally, we discussed about our software architecture which is based on ROS and described the basic control packages and nodes running on different layers.
\vspace{-0.1em}
\bibliographystyle{IEEEtran}
\bibliography{output.bbl}
\end{document}